# Unsupervised Non Linear Dimensionality Reduction Machine Learning methods applied to Multiparametric MRI in cerebral ischemia: Preliminary Results

Vishwa S. Parekh[1,3], Jeremy R. Jacobs[1], Michael A. Jacobs[1,2]

[1] The Russell H. Morgan Department of Radiology and Radiological Sciences and [2] Sidney Kimmel Comprehensive Cancer Center.

The Johns Hopkins University School of Medicine, Baltimore, MD 21205, USA

[3]Department of Computer Science, The Johns Hopkins University, Baltimore, MD 21208

## Abstract

The evaluation and treatment of acute cerebral ischemia requires a technique that can determine the total area of tissue at risk for infarction using diagnostic magnetic resonance imaging (MRI) sequences. Typical MRI data sets consist of T1- and T2-weighted imaging (T1WI, T2WI) along with advanced MRI parameters of diffusion-weighted imaging (DWI) and perfusion weighted imaging (PWI) methods.  Each of these parameters has distinct radiological-pathological meaning. For example, DWI interrogates the movement of water in the tissue and PWI gives an estimate of the blood flow, both are critical measures during the evolution of stroke. In order to integrate these data and give an estimate of the tissue at risk or damaged; we have developed advanced machine learning methods based on unsupervised non-linear dimensionality reduction (NLDR) techniques. NLDR methods are a class of algorithms that uses mathematically defined manifolds for statistical sampling of multidimensional classes to generate a discrimination rule of guaranteed statistical accuracy and they can generate a two- or three-dimensional map, which represents the prominent structures of the data and provides an embedded image of meaningful low-dimensional structures hidden in their high-dimensional observations. In this manuscript, we develop NLDR methods on high dimensional MRI data sets of preclinical animals and clinical patients with stroke. On analyzing the performance of these methods, we observed that there was a high of similarity between multiparametric embedded images from NLDR  methods  and the ADC map and perfusion map. It was also observed that embedded scattergram of  abnormal (infarcted or at risk)  tissue can be visualized and provides a mechanism for automatic methods to delineate potential stroke volumes and early tissue at risk.

## 1.   INTRODUCTION

The evaluation and treatment of acute cerebral ischemia have been revolutionized by the advent of thrombolytics and functional magnetic resonance imaging (MRI) techniques.   Therapeutic intervention using thrombolytic therapy can result in the restoration of blood flow potentially result in salvaging ischemic brain tissue that is not already infarcted[1-3].   Moreover, the introduction of perfusion- and diffusion -weighted MRI imaging (PWI and DWI, respectively) allows for the identification of the tissue at risk[4-7].  Moreover, by using multiparametric MRI data, the onset of the ischemic event can be determined using the Apparent Diffusion Coefficient (ADC) map and the T2-weighted image[6].  Current stroke MRI protocols consist of T1- and T2-weighted imaging (T1WI, T2WI) along with DWI and PWI.  Each of these MRI parameters have distinct radiological-pathological meaning.  For example, DWI interrogates the movement of water in the tissue and perfusion gives an estimate of the blood flow, both of these are critical measures during stroke and provide a way to gauge if the tissue is a risk for infraction or has infracted.  If thrombolytics have been administrated, a question is has the tissue been  reprefused?  Thus, methods are needed to integrate all of the imaging data and give an estimate of the tissue at risk, so that appropriate clinical decisions can be determined.    We have developed hybrid unsupervised advanced machine learning methods based on nonlinear dimensionality reduction (NLDR) techniques to quantitatively integrate multiparametric radiological images into an embedded image[8].  Then we apply a clustering algorithm to define tissue at risk versus tissue infarcted. The NLDR methods are a class of algorithms that uses mathematically defined manifolds for statistical sampling of multidimensional classes to generate a two- or three-dimensional map, which represents the prominent structures of the data and provides an embedded image of meaningful low-dimensional structures hidden in their high-dimensional observations.  In this manuscript, we developed a hybrid NLDR method with a clustering classification and compare performance of this new method on high dimensional MRI data sets applied to a group of preclinical animals and clinical patients with stroke.

## 2. METHODS AND MATERIALS

**2.1 Theory**

**2.1.1 Manifold Learning of Cerebral Ischemia: Nonlinear Dimensionality Reduction(NLDR)**

We hypothesize that different tissue classes in brain such as, white matter, gray matter, CSF, and abnormal tissue will exhibit distinct characteristic MRI signal intensities and that these different MR signals can provide a way to classify the tissue [9, 10].  Therefore, to classify ischemia in tissue, a novel model of tissue signatures was developed based on segmentation of abnormal tissue from normal tissue by utilizing NLDR methods.  NLDR methods are mathematical descriptors of manifolds and can be used  to identify and/or classify tissue types in n-dimensional feature space created from the multiparametric MRI data as described below.

Typical preclinical MRI data used in this study are shown in Fig. 1.  We developed modified versions of Isomap, diffusion maps (DfM) and LLE to construct NLDR embedded mappings to define different tissues in multiparametric stroke MRI data [11-13].  To that end, we mapped high-dimensional MRI data to lower dimensions, while maintaining and preserving the most important features. Our new machine-learning NLDR method is used to detect the intrinsic dimensionality of the high-dimensional data that can represent the structure of the data. Mathematically, a data set $X \subset R^{D(images)}$, that is $X \subset R^{D(x_1, x_2, \ldots x_n) = D(Images)}$ where, $x_1, x_2, \ldots, x_n$ = T1WI,T2WI, DWI, perfusion MRI or other images, that have intrinsic dimensionality d<D, if X can be defined by d points or parameters that lie on a manifold.  By definition, a manifold is a topological space that is locally Euclidean, i.e., around every point, there exists a neighborhood that is topologically the same as the open unit ball in Euclidian space (that is compact) [14]. Thus, the manifold learning method is used to determine points or locations within dataset X (e.g., DWI, PWI, etc.) lying on or near a manifold with intrinsic (lower) dimensionality d and is embedded in the higher dimensional space (D). Therefore, NLDR methods map dataset $X=\{x_1, x_2, \ldots, x_n\} \subset R^{D(images)}$ into a new dataset $Y=\{y_1, y_2, \ldots, y_n\} \subset R^d$ with dimensionality d, while retaining the geometry of the data as much as possible. NLDR techniques have the ability to deal with complex nonlinear data.

**Isomap:** If the high-dimensional data lies on or near a curved manifold, Euclidean distance does not take into account the distribution of the neighboring data points and might consider two data points as near points, whereas their distance over the manifold is much larger than the typical inter-point distance. Isomap overcomes this problem by preserving pairwise geodesic (or curvilinear) distances between data points[12]. The geodesic distance (GD) is the distance between two points measured over the manifold. GDs between the data points could be computed by constructing a neighborhood graph G (every data point xi is connected with its k nearest neighbors, xij). GDs are estimated using Dijkstra's shortest-path algorithm to find the shortest path between two points in the graph. GDs between all data points form a pairwise GD matrix. The low-dimensional space Y is computed then by applying multidimensional scaling (MDS) while retaining the GD pairwise distances between the data points[15].

**Diffusion maps (DfM):** Diffusion maps find the subspace that best preserves the diffusion interpoint distances by defining a Markov random walk on a graph.  Pairs of data points with a high forward transition probability have a small diffusion distance and those with high density have more weight. Based on spectral theory on the random walk, the low-dimensional representation Y can be obtained using the d nontrivial eigenvectors of the distance matrix D: As the graph is fully connected, eigenvector v1 of the largest eigenvalue is discarded and the eigenvectors are normalized by their corresponding eigenvalues[13].

**Locally Linear Embedding (LLE):** In contrast to Isomap, LLE preserves the local properties of the data, which allows for successful embedding of nonconvex manifolds via conformal mapping[11]. LLE assumes that the manifold is locally linear and constructs the local properties of the data manifold using a weighted summation of the k nearest neighbors for each point. Then it fits a hyperplane through the data point $x_i$ and its k nearest neighbors $x_{ij}$. The reconstruction weights $W_i$ are invariant to translation, rotation, and rescaling to imply the local linearity assumption. Thus, any linear mapping of the hyperplane to a space of lower dimensionality preserves the reconstruction weights in the space of lower dimensionality.

**2.1.2 Consensus Clustering**

Manifold Learning of cerebral ischemia as described in Section 2.1.1 produces a resultant embedded image, in which we hypothesize contains important information of the data structure. In order to characterize and classify the different tissue types into different classes, we applied a modified version consensus clustering on the embedded image. Consensus Clustering is an unsupervised method based on K-means, with additional steps of cluster size and number adjustment [16, 17]. The algorithm is based on development of a pixel similarity matrix by running K-means algorithm at different values of K and with different cluster center initializations and then form an optimum clustering using this similarity matrix.

The algorithm is summarized by the following steps:

- Go from k = k1 to k = k2
    - Do H times (defined by the user)
        - Randomly select k cluster centers and run K means algorithm on this initialization
        - Update the similarity matrix:
            - for each pair (i,j) in the same cluster C, similarity(i,j) = similarity(i,j) + 1
- Normalize the similarity matrix
- Find majority voting be setting a threshold t, on the similarity matrix such that if similarity(i,j) > t, add them to the same cluster, and if they belong to different clusters, merge the clusters.

**2.1.2 Tissue Characterization**

In our hybrid NLDR method, the embedded image provides a visualization of all integrated information from the MRI parameters into a single image. Then using consensus clustering allows for automatic segmentation of this low dimensional embedded image for tissue characterization of the data into different tissue classes. The MRI parameters we used for constructing embedded image were the most biological revelant, the perfusion, ADC and T2 map for all the time points. The imput NLDR parameters were set as follows: neighborhood parameter k for the Isomap and LLE was varied between 20 and 80. The sigma criteria for diffusion maps was varied between 100 and 60. After NLDR was applies, the consensus clustering was run. We varied the H parameter between 3 and 13 to determine the optimal clusters. The minimum number of clusters was set to three, since, there are three normal tissue classes, white and gray matter with CSF. The maximum number of clusters was set to 13 in order to have a maximum of 10 different classes of potential abnormal (lesion) tissue based on literature values [9]. The threshold value for clustering the data from the similarity matrix was set to 0.75 for one and 0.80 for another set of observations. Using these thresholds, a final tissue cluster was obtained. From the tissue cluster, the labels of interest representing the abnormal tissues were identified.

### 2.2 Preclinical Data

All studies were performed in accordance with the institutional guidelines for animal research under a protocol approved by the Institutional Care of Experimental Animals Committee. Male Wistar rats (n=19) weighing 270 -310 g were utilized in these experiments separated into three different groups at different time points ranging from acute (4-8h, n=6), subacute (16-24h, n=8) and chronic (48-168h, n=5) and subjected to permanent middle cerebral artery occlusion(MCA) by a method of intraluminal vascular occlusion[18]. This method provides a focal infarct in the striatum, i.e., caudate putamen and globus pallidus, that often extends into the cortex over time.

#### 2.2.1 Preclinical MRI Measurements

After MCA occlusion, the animal was placed in the magnet and MRI data sets were acquired on a 7T, 20cm bore, superconducting magnet (Magnex Scientific Inc.) interfaced to a SMIS (Surry Medical Imaging Systems, Inc.) console. The animal was fixed using stereotaxic ear bars to reduce motion during the experiment. During MRI measurements, anesthesia was maintained using 3.5% halothane, and maintained with 0.75-1.0% halothane in 70% N20 and 30% O2 gas mixture. Rectal temperature was monitored and controlled using a feedback-controlled water bath. PWI, DWI, and T2WI were performed on each group of animals after MCA occlusion at acute, subacute, and chronic time points. Perfusion measurements were obtained using an arterial spin tagging technique [19]. Diffusion weighted images (DWI) were obtained using the method multislice spin-echo sequence (7 slices, 128x128 matrix, field of view (FOV)=3.2 cm and TR/TE 1500/40 ms, NEX=2) with diffusion

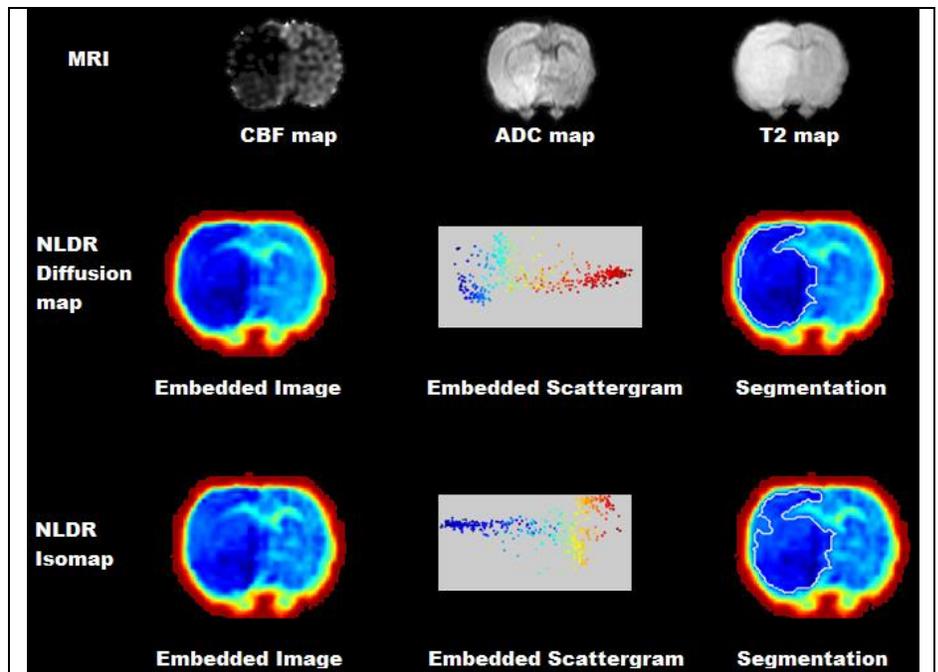

**Figure 1**. **Top Row**) Typical MRI input parameters consisting of the cerebral blood flow maps, Adc map, and T2 weighted image at 168 hrs post ictus. **Middle Row)** Resultant embedded images and scattergram from the Diffusion mapping. **Bottom Row**) Similar results from the application of the ISOMAP algorithm demonstrate excellent delineation of the ischemic region and normal tissue

weighted gradient b-values of 0, 200, 400, 600, and 800 s/mm2 T2 weighted images were acquired in a multislice (7 contiguous slices), multiecho sequence (FOV=3.2cm, TR /TE = 3000 / 30, 60, 90, and 120 ms) using a 128 X128 image matrix and 2 mm slice thickness. After each time point, the animals were sacrificed and the brain tissue was processed to histological analysis.

### 2.3 Clinical Data

All studies were performed in accordance with the institutional guidelines for clinical research under a protocol approved by the Institutional Review Board (IRB) and all HIPPA agreements were followed. All patients (n=5, (10 total studies),age =71.6±17.3 yrs) with ischemic stroke underwent an acute stroke MRI protocol at different time points divided into acute (0-12hr) and subacute (24-168 hr) time points. The acute stroke protocol consisted of sagittal T1, axial spin echo T2WI) pre- and post-gadolinium T1WI perfusion, DWI, and circle of Willis three dimensional phase contrast magnetic resonance angiography (MRA). All MRI were performed on a 3T Siemens MRI unit using a phased array head coil. The MRI parameters were: T1WI sagittal image (TR/TE 200/2.46ms, field of view (FOV)=240x240, slice thickness (ST) = 5mm),

axial T2WI Flair (TR/TE/TI = 9000/105/2500ms, FOV=230x230, ST=4) and T1WI perfusion (TR/TE/1350/30ms, FOV=230x230, ST=4)., Axial DWI (TR/TE=9000/98ms, b-values=1000, and 0 s/mm$^2$, FOV 230x230, matrix=128x128, ST=4 mm) were obtained.

### 2.4 MRI Image Preprocessing and Analysis
Maps of ADC were created for each time point using a least squares fit from slope of the signal intensity from the DWI on a pixel by pixel basis from the equation

$$ADC = \frac{\ln(\frac{S_i}{S_0})}{b} \quad (1)$$

Where b = the diffusion gradient values, $S_0$ = 1$^{st}$ image (b=0), and $S_i$ = ith image. The T2 relaxation maps were generated using the equation

$$T2 = \frac{\ln(\frac{S_i}{S_0})}{TE} \quad (2)$$

Where TE = the echo time, $S_0$ = 1$^{st}$ image (b=0), and $S_i$ = ith image. The cerebral blood flow maps (CBF) and perfusion time to peak (TTP) maps were calculated from T1WI perfusion data. **Figure 1** demonstrates a typical MRI sequences from a preclinical data set with corresponding embedded images and scattergrams.

### 2.5 Statistical Methods
Summary statistics (mean and standard deviations) were calculated for each of the preclinical and clinical MRI parameters. Regression and nonparametric Spearman rank correlations were performed between the "radiological" gold standard and the hybrid NLDR segmented and classified areas. Statistical significance was sent at p<0.05.

## 3. RESULTS
The hybrid multiparametric manifold learning based segmentation and clustering method identified different regions of tissue damage at each time point after stroke in both the preclinical and clinical setting. **Figure 2** demonstrates the use of NLDR on an animal that underwent induction of cerebral ischemia at 4 hours. The NLDR methods were able to segment different degrees of tissue damage and then the application of the clustering algorithm allowed for classification of the tissue types into "tissue at risk" versus "infracted" tissue. The tissue at risk was characterized by intermediate ADC map values (0.646x10$^{-3}$ mm$^2$/s) and slightly reduced CBF (125 mL/g/s) compared to contralateral ADC and CBF map values (0.750x10$^{-3}$ mm$^2$/s and 163.4 mL/g/s, respectively). There were slight changes in the T2 map values in the same regions (38 vs 34ms). However, compared to "infracted" tissue, the ADC and CBF map values were lower (0.358x10$^{-3}$ mm$^2$/s and 14.1 mL/g/s) with an increase in the T2 map value (43ms). The tissue classification in the consensus clustering was uniform across each of the NLDR images. The regions colored in yellow depict the infarcted region and red colored regions depict "tissue at risk". The DWI, T2, CBF lesion areas compared to the predicted NLDR infarcted and predicted total tissue affected (infarcted and tissue at risk) for the preclinical animals are summarized in **Table 1**. Indeed, there was high correlation between the DWI and CBF areas compared to Dfm (R=0.7 and 0.77) and Isomap (R=0.8 and 0.7), but the correlations were low for LLE (R=0.2 and 0.54) at the acute (0-8hrs) time point. There was low correlation between the T2 map areas and all NLDR methods as expected. At the subacute time point, the correlation between the T2 map areas improved best with Isomap (R=0.98). Finally, at the chronic time point, all NLDR method defined lesion areas were highly correlated with the T2 map area (Dfm, R=0.71, Isomap, R=0.83, and LLE, R=0.40). **Figure 3** plots the predicted areas using Diffusion maps in comparison to the T2, DWI and CBF lesions at the acute stage for the preclinical study data.

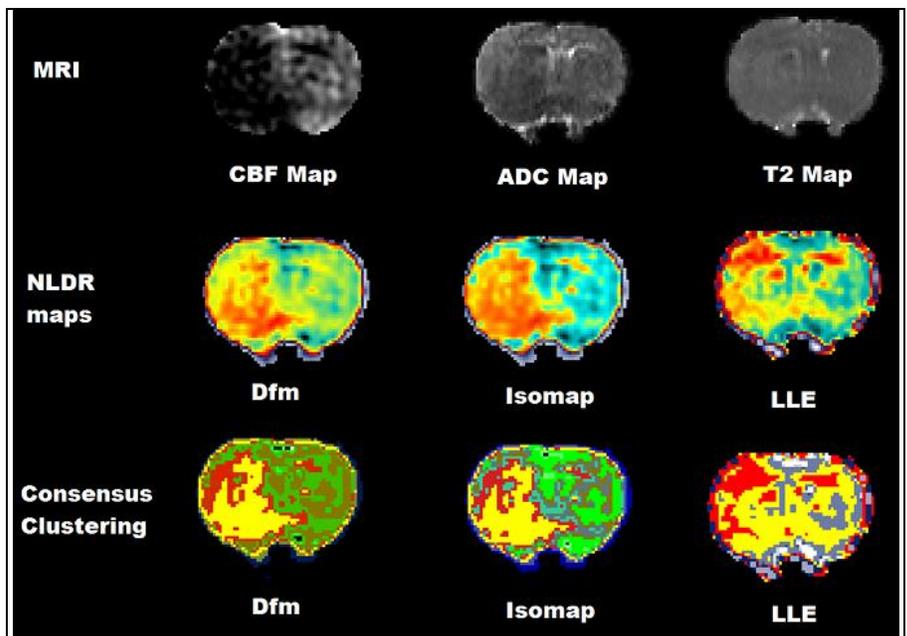

Figure 2: Illustration of multiparametric MRI data, the resultant embedded images from NLDR methods and corresponding segmentations, yellow (for infarcted) and red (for mismatch) regions, using consensus clustering algorithm in the representative animal at acute time point.

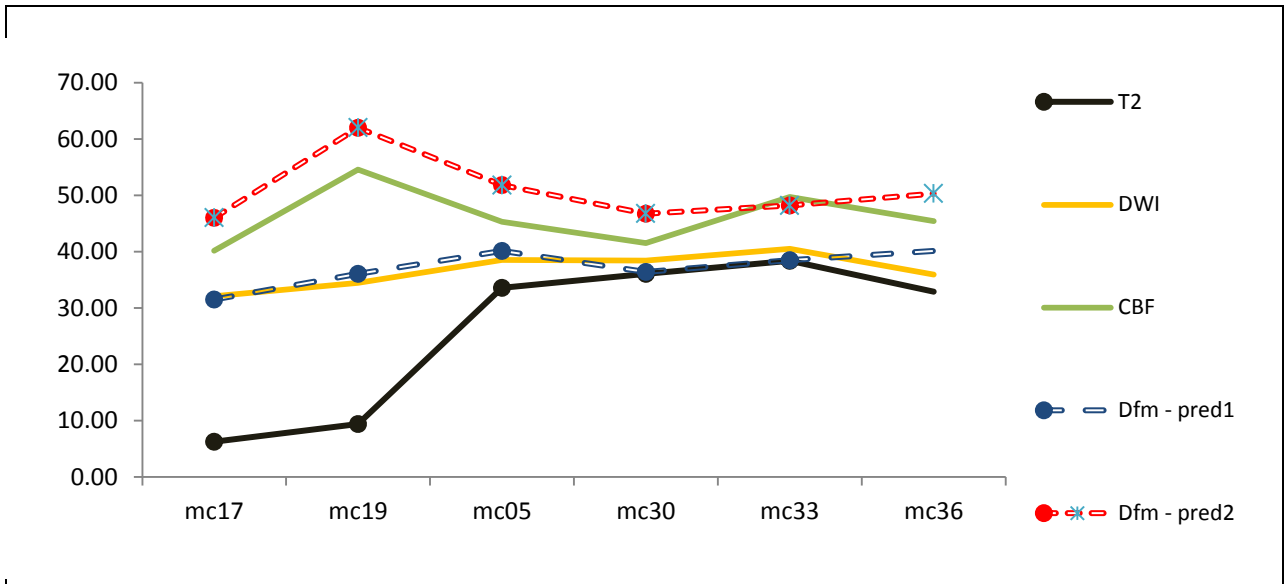

Figure 3. Comparative analysis of stroke lesion areas obtained from T2WI, DWI, CBF and DIffusion Map(Dfm). Dfm pred1 represents predicted classes of infarcted tissue. Whereas, the Dfm pred2 represents predicted volume of complete affected tissue, ie, combined tissue at risk and infarcted tissue.

Table 1. PreClinical Stroke data: Lesion areas defined by T2-weighted imaging (T2WI), Diffusion weighted imaging (DWI), Perfusion imaging of the cerebral blood flow (CBF) and NLDR methods.

| Animal/Areas (mm²) | T2WI | DWI | CBF | Dfm Infarcted | Dfm Total Affected | Isomap Infarcted | Isomap Total Affected | LLE Infarcted | LLE Total Affected |
|---|---|---|---|---|---|---|---|---|---|
| **Acute (4-8hrs)** | | | | | | | | | |
| **mc17** | 6.25 | 32.13 | 40.19 | 31.50 | 46.00 | 31.13 | 45.63 | 36.00 | 46.06 |
| **mc19** | 9.38 | 34.50 | 54.56 | 36.06 | 62.00 | 34.00 | 54.13 | 35.69 | 54.56 |
| **mc05** | 33.56 | 38.56 | 45.31 | 40.13 | 51.81 | 42.25 | 50.69 | 34.00 | 52.19 |
| **mc30** | 36.06 | 38.44 | 41.56 | 36.38 | 46.75 | 34.13 | 47.38 | 40.31 | 44.75 |
| **mc33** | 38.38 | 40.50 | 49.75 | 38.50 | 48.19 | 41.56 | 46.00 | 36.31 | 60.88 |
| **mc36** | 32.88 | 35.94 | 45.44 | 40.13 | 50.31 | 37.75 | 49.69 | 31.88 | 31.88 |
| **Subacute (12-14 hrs)** | | | | | | | | | |
| **mc13** | 40.00 | 38.56 | 50.75 | 33.94 | 50.06 | 40.38 | 60.88 | 20.25 | 61.13 |
| **mc14** | 31.81 | 27.56 | 38.38 | 32.13 | 42.31 | 30.13 | 44.19 | 32.31 | 37.31 |
| **mc16** | 48.06 | 42.18 | 49.13 | 47.56 | 52.94 | 49.81 | 55.93 | 38.88 | 55.81 |
| **mc 27** | 34.50 | 26.25 | 41.88 | 38.56 | 45.81 | 30.88 | 45.25 | 31.69 | 51.06 |
| **mc04** | 26.94 | 27.87 | 36.75 | 55.94 | 55.94 | 28.44 | 39.63 | 43.31 | 54.31 |
| **mc21** | 27.88 | 26.37 | 36.38 | 28.25 | 43.75 | 25.38 | 41.88 | 65.44 | 64.94 |
| **mc38** | 39.25 | 37.38 | 51.94 | 39.38 | 67.81 | 38.75 | 51.06 | 36.75 | 49.00 |
| **Chronic (48-168hrs)** | | | | | | | | | |
| **mc11** | 45.56 | 41.81 | 42.63 | 43.25 | 43.25 | 44.44 | 44.44 | 36.19 | 36.19 |
| **mc18** | 58.25 | 54.63 | 57.75 | 62.88 | 62.88 | 58.75 | 58.75 | 58.81 | 58.81 |

| | | | | | | | | |
|---|---|---|---|---|---|---|---|---|
| **mc24** | 56.38 | 46.50 | 54.13 | 65.56 | 65.56 | 65.19 | 65.19 | 86 | 52.06 |
| **mc03** | 40.56 | 33.44 | 38.44 | 32.06 | 32.06 | 31.56 | 31.56 | 32.38 | 32.38 |
| **mc08** | 50.31 | 40.19 | 48.50 | 29.75 | 29.75 | 43.25 | 43.25 | 36.69 | 36.69 |

The clinical results using the hybrid method is demonstrated in Figure 4 on a patient left hemispheric stoke within 12 hrs. The TTP map demonstrates the entire MCA area is at risk (Note: the higher intensity on the TTP maps indicates prolonged perfusion time compared to normal tissue). But, densely ischemic tissue is shown by the DWI and ADC map around the left basal ganglia extending from the inferior portion of the caudate head. There a "perfusion-diffusion" mismatch. The ADC map values in the lesion areas were significantly different from the contralateral side (0.379± 0.131 x10$^{-3}$mm$^2$/s versus 0.690±0.11 x10$^{-3}$ mm$^2$/s). Similar the TTP values were different. The Isomap was able to define all the tissue at risk. Figure 5 shows the progression of ischemic tissue after 24-48 hrs in the same patient in Figure 4. The ADC and TTP maps values increased, but remained lower than the contralateral hemisphere. Table 2 summarizes the NLDR methods on the clinical studies. The segmented ischemic tissue regions had significant correlation with MRI defined areas.

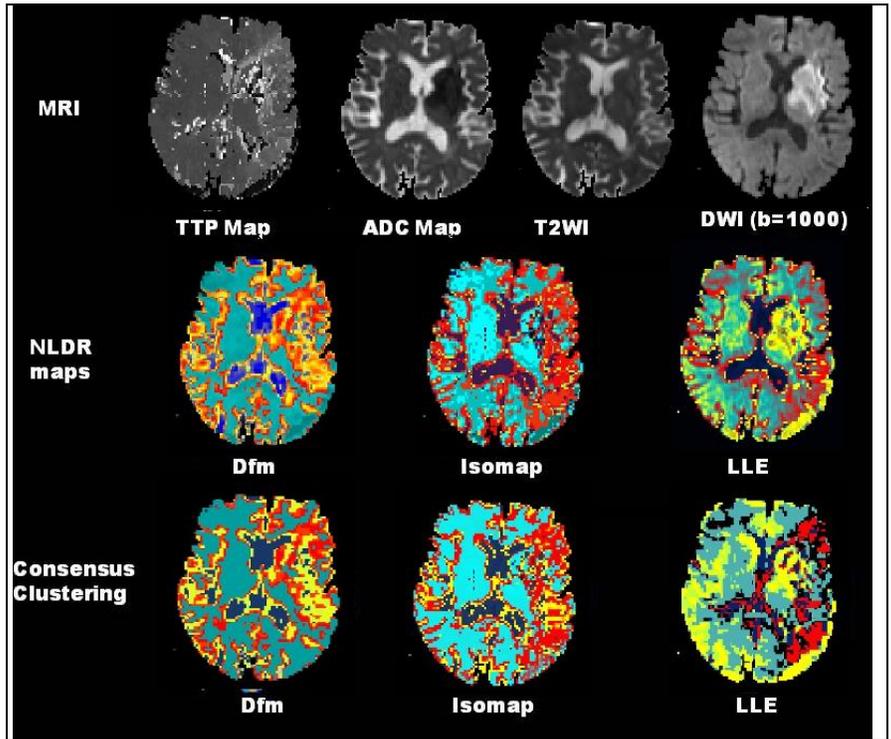

Figure 4: Demonstration of the hybrid NLDR methods on an acute clinical case (Patient 1. 86 y/o female) with left middle cerebral artery (MCA) territory infarct. **Top Row)** The TTP map demonstrates the entire MCA area is at risk with the ADC map showing an insular infarct area. There is a clear "perfusion-diffusion" mismatch. This is demonstrated in the **Middle and Bottom Rows)** by the NLDR methods and clustering algorithms. The Isomap demonstrates the entire area at risk.

**Table 2. Clinical Stroke Data: Volumes defined by T2WI, DWI, CBF and NLDR methods.**

| Patient/Areas | T2 (mm$^2$) | DWI (mm$^2$) | CBF (TTP map) (mm$^2$) | Dfm Infarcted (mm$^2$) | Dfm Total (mm$^2$) | Isomap Infarcted (mm$^2$) | Isomap Total (mm$^2$) | LLE Infarcted (mm$^2$) | LLE Total (mm$^2$) |
|---|---|---|---|---|---|---|---|---|---|
| **Patient1: <08 hrs** | 4.33 | 10.99 | 37.62 | 19.2 | 45.3 | 20.82 | 56.79 | 18.54 | 59.1 |
| **Patient1: 24 hrs** | 22.46 | 38.96 | 29.38 | 47.82 | 33.42 | 45.09 | 45.9 | 30.15 | 9.27 |
| **Patient2: <08 hrs** | 0.00 | 3.84 | 30.50 | 6.69 | 33.84 | 4.2 | 25.2 | 7.92 | 33.21 |
| **Patient2: 24 hrs** | 7.43 | 5.03 | 9.16 | 19.8 | 19.8 | 7.29 | 7.29 | 15.33 | 15.33 |

## 4. DISCUSSION AND CONCLUSION

In this paper, we developed a novel hybrid scheme to integrate multiple brain MRI data into a single embedded image, which carries information from each of input images using advanced machine learning methods, in particular, modified NLDR and clustering techniques. The resulting embedded image was able to visualize and segment infracted and tissue at risk in stroke from the adjacent normal white and gray matter with excellent correlation to radiological gold standards.

The NLDR methods, in general, were able to segment and visualize the underlying structure of the normal brain tissue, infracted and tissue at risk. Why this is important is that the NLDR methods could be used to gauge how much tissue is potentially salvable and amendable to therapeutic interventions [2, 20]. Using multiparametric stroke data represent complex high dimensional data and not one single parameter conveys all the necessary information to determine the tissue at risk [6]. Integration of the stroke MRI parameters is needed and our methods provide an opportunity to integrate them. Our results demonstrate that when the NLDR methods are applied to the stroke MRI data, we were able segment the lesion and provided excellent visualization of the tissue structure. Furthermore, this mapping of high dimensional data to a lower dimension provides a mechanism to explore the underlying contribution of each MR parameter to the final output image. Our results are similar to other reports using NLDR in biomedical applications [8, 21]. With the use of NLDR methods, we selected the most important biophysical parameters to represent the varying degree of tissue response to cerebral ischemia. Our data suggest the Isomap gave the best results of densely ischemic tissue compared to the final area in ADC map and the lesion area of the perfusion map. The ISOMAP was able to segment out the area of perfusion-mismatch. Moreover, the embedded scattergram will provide a quantitative relationship between the pixels that identify various tissues with the normal and infarcted regions. There was a high degree of similarity between multiparametric embedded and the ADC map and perfusion map. Using the hybrid unsupervised NLDR methods, we have demonstrated that embedded scattergram of abnormal (infarcted or at risk) tissue can be visualized and provides a mechanism for automatic methods to delineate potential stroke volumes and early tissue at risk. Thus, the NLDR methods show a lot of promise in working with multiparametric MRI data in cerebral ischemia.

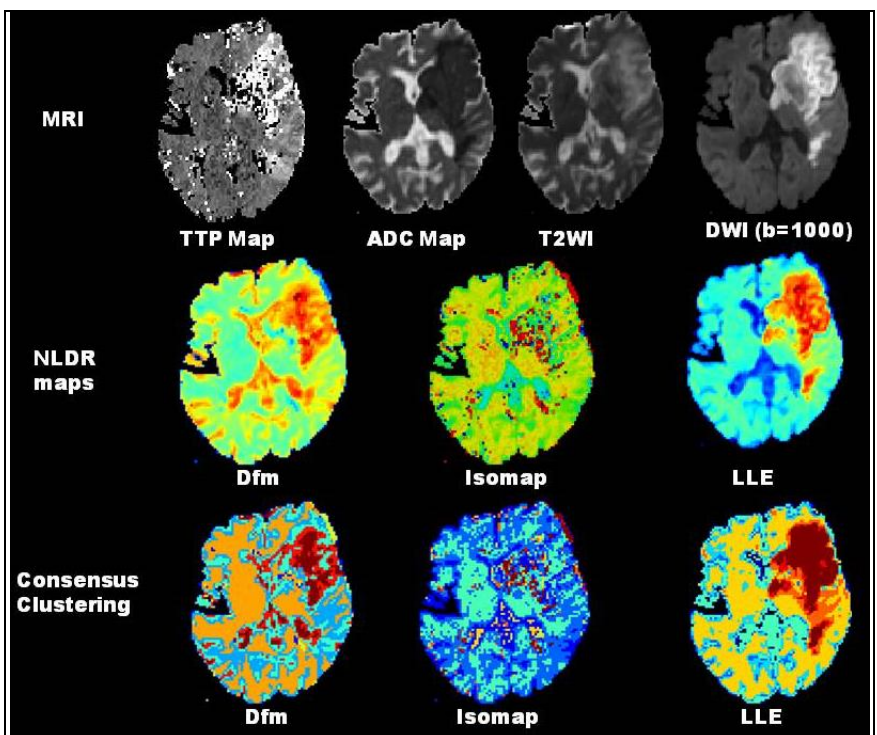

Figure 5. Demonstrate the spread of ischemic tissue in a stroke patent after 24-48 hrs. **Top Row)** multiparametric MRI data: TTP map, ADC map, T2 and DWI images. **Middle and Bottom Row)** the resultant embedded images from NLDR methods and corresponding segmentations, regions, using consensus clustering algorithm.

## 4. Acknowledgment

We thank Drs. Richard Leigh, Peter Barker, and Michael Chopp for supporting this work. This work was supported in part by National Institutes of Health grant numbers: P50CA103175, 5P30CA06973, U01CA070095, and U01CA140204.